\documentclass{article}
\usepackage{spconf,amsmath,graphicx}
\usepackage{cite}
\usepackage{amsmath,amssymb,amsfonts,bm}
\usepackage{algorithmic}
\usepackage{graphicx}
\usepackage{textcomp}
\usepackage{xcolor}
\usepackage{float}
\usepackage{algorithm}
\usepackage{setspace}
\usepackage[colorlinks]{hyperref}
\usepackage[utf8x]{inputenc}

\title{3D Point Cloud Denoising via Deep Neural Network based Local Surface Estimation}
\name{Chaojing Duan$\,^{1}$ \qquad Siheng Chen$\,^{2}$ \qquad Jelena Kova\v cevi\' c$\, ^{3}$}

\address{$^{1}$ Department of Electrical \& Computer Engineering, Carnegie Mellon University, Pittsburgh, USA\\
    $^{2}$ Mitsubishi Electric Research Laboratories, Boston, USA\\
    $^{3}$ Tandon School of Engineering, New York University, Brooklyn, USA}

\begin{document}
%
\maketitle
\begin{abstract}
We present a neural-network-based architecture for 3D point cloud denoising called neural projection denoising (NPD). In our previous work, we proposed a two-stage denoising algorithm, which first estimates reference planes and follows by projecting noisy points to estimated reference planes. Since the estimated reference planes are inevitably noisy, multi-projection is applied to stabilize the denoising performance. NPD algorithm uses a neural network to estimate reference planes for points in noisy point clouds. With more accurate estimations of reference planes, we are able to achieve better denoising performances with only one-time projection. To the best of our knowledge, NPD is the first work to denoise 3D point clouds with deep learning techniques. To conduct the experiments, we sample 40000 point clouds from the 3D data in ShapeNet to train a network and sample 350 point clouds from the 3D data in ModelNet10 to test. Experimental results show that our algorithm can estimate normal vectors of points in noisy point clouds. Comparing to five competitive methods, the proposed algorithm achieves better denoising performance and produces much smaller variances. Our code is available at \url{https://github.com/chaojingduan/Neural-Projection}.
\end{abstract}
\vspace{-1mm}
\section{Introduction}
\label{intro}
The rapid development of 3D sensing techniques and the emerging field of 2D image-based 3D reconstruction make it possible to sample or generate millions of 3D points from surfaces of objects~\cite{pcl, pc_tutorial,sample_surface}. 3D point clouds are discrete representations of continuous surfaces and are widely used in robotics, virtual reality, and computer-aided shape design. 3D point clouds sampled by 3D scanners are generally noisy due to measurement noise, especially around edges and corners~\cite{sensor}. 3D point clouds reconstructed from multi-view images contain noise since the reconstructed algorithms fail to manage matching ambiguities~\cite{image_pc, machine_noise}. The inevitable noise in 3D point clouds undermines the performance of surface reconstruction algorithms and impairs further geometry processing tasks since the fine details are lost and the underlying manifold structures are prone to be deformed~\cite{siheng}. 

However, 3D point clouds denoising or processing is challenging because 3D point clouds are permutation invariant and the neighboring points representing the local topology interact without any explicit connecting information. To denoise 3D point clouds, we aim to estimate the continuous surface localized around each 3D point and remove noise by projecting noisy points to the corresponding local surfaces. The intuition is that noiseless points are sampled from surfaces. To estimate local surfaces, we parameterize them by 2D reference planes. By projecting noisy points to the estimated reference planes, we ensure that all the denoised points come from the underlying surfaces.

Deep neural networks have shown ground-breaking performances in various domains, such as speech processing and image processing~\cite{dl_language}. Recently, several deep-learning architectures have been proposed to deal with 3D point clouds in tasks such as classification, segmentation, and upsampling~\cite{2016pointnet, folding, upsampling}. In this work, we learn reference planes from noisy point clouds and further reduce noise with deep learning; we name the proposed algorithm as neural projection denoising (NPD). Estimated reference planes are represented by normal vectors and interceptions. The reason for using deep learning is that previous algorithms are not robust enough to noise intensity, sampling density, and curvature variations. These algorithms need to define neighboring points to capture local structures. However, it is difficult to choose the neighboring points adaptively according to the sampling density or curvature variation. 

In our experiments, the point clouds used for training are sampled from the 3D dataset ShapeNet and the point clouds used for testing are sampled from the 3D dataset ModelNet10~\cite{shapenet, modelnettest}. The experimental results show that NPD outperforms four of five other denoising algorithms in all seven categories and achieves the lowest variance in both evaluation metrics. 

\textbf{Contributions.} 1) To the best of our knowledge, NPD is the first to directly deal with 3D point clouds for denoising tasks with deep learning techniques; 2) NPD can estimate normal vector for each point with both local and global information and is less affected by noise intensity and curvature variation; 3) NPD can denoise noisy point clouds without defining neighboring points for noisy point clouds or calculating the eigendecomposition to estimate local geometries; 4) NPD provides the possibility of 3D point cloud parameterization with the combination of local and global information. 

\vspace{-3mm}
\section{Related work}
\textbf{Point cloud denoising.}
3D point cloud denoising has been tackled by various approaches: mesh-based denoising, graph-based denoising, and projection-based denoising. Bilateral filter (BF) and partial differential equations (PDE) are widely used in mesh and point cloud denoising~\cite{denoise_bf, denoise_pde1}, but these mesh-based algorithms cause shrinkage and deformation~\cite{denoise_nld}. Graph-based denoising (GBD) algorithms have received increasing attention because the Laplace-Beltrami operator of manifolds can be approximated by the graph Laplacian~\cite{graph_manifold, newgraph}; however, a graph constructed from a noisy point cloud is also noisy and cannot reflect the true manifold, causing deformation issues~\cite{graph_analysis}. Projection-based denoising algorithm named weighted multi-projection (WMP) in~\cite{WMP} estimates reference planes for 3D points and project noisy 3D points to the planes multiple times for denoising purpose. NPD estimates the reference planes with deep learning techniques and project noisy point clouds only once.

\textbf{Deep learning on 3D data.}
Researchers first attempted to handle 3D data with deep learning techniques by voxelizing 3D shapes and applying 3D convolutional neural networks~\cite{modelnettest}. Voxelization as a 3D data pre-processing procedure is computationally intensive and leads in quantization artifacts~\cite{2016pointnet}. The authors in~\cite{2016pointnet} proposed an architecture that directly consumed 3D point clouds without voxelizing or converting for classification and segmentation tasks. In the paper, we adopt the idea and redesign the framework to estimate normal vectors of points in a noisy point cloud and directly denoise 3D point cloud via deep learning techniques.  

\vspace{-3mm}
\section{3D point cloud denoising algorithm}
\textbf{Problem formulation.}
Let $\mathcal{S} = \{\mathbf{p}_i \in \mathbb{R}^3 \;|\: i = 1, ..., N\}$ be a 3D noiseless point cloud, where $N$ is the total number of points in $\mathcal{S}$, and each point $\mathbf{p}_i = [x_i, \, y_i, \, z_i]^T$ is a coordinate vector. Let $\widetilde{\mathcal{S}} = \{\widetilde{\mathbf{p}}_i \in \mathbb{R}^3 \;|\; i = 1, ..., N\}$ represent a noisy 3D point cloud, and each point $\widetilde{\mathbf{p}}_i = [\widetilde{x_i}, \, \widetilde{y_i}, \, \widetilde{z_i}]^T$ is a coordinate vector. Note that $\widetilde{\mathbf{p}}_i = \mathbf{p}_i + \mathbf{n}_i$, where $\mathbf{n}_i$ is a 3D noise vector attached to the point $\mathbf{p}_i$ and $\mathbf{n}_i \mathtt{\sim} \mathcal{N}(\mathbf{0} , \bm{\Sigma})$. 

Since 3D point clouds are sampled from surfaces, they are essentially 2D manifolds embedded in 3D space. These surfaces can be locally approximated by 2D reference planes. One of our goals is learning reference planes for points with the existence of noise, but the accurate reference planes for the ground-truth surface are unknown. Unlike the work in~\cite{WMP} which calculate the reference planes by constructing weighted covariance matrices from noisy point clouds, NPD estimates the reference planes by learning the global and local geometries via deep learning techniques. 
\begin{figure}[htb!]
\centerline{\includegraphics[width=3.2in , height = 1.2in, keepaspectratio]{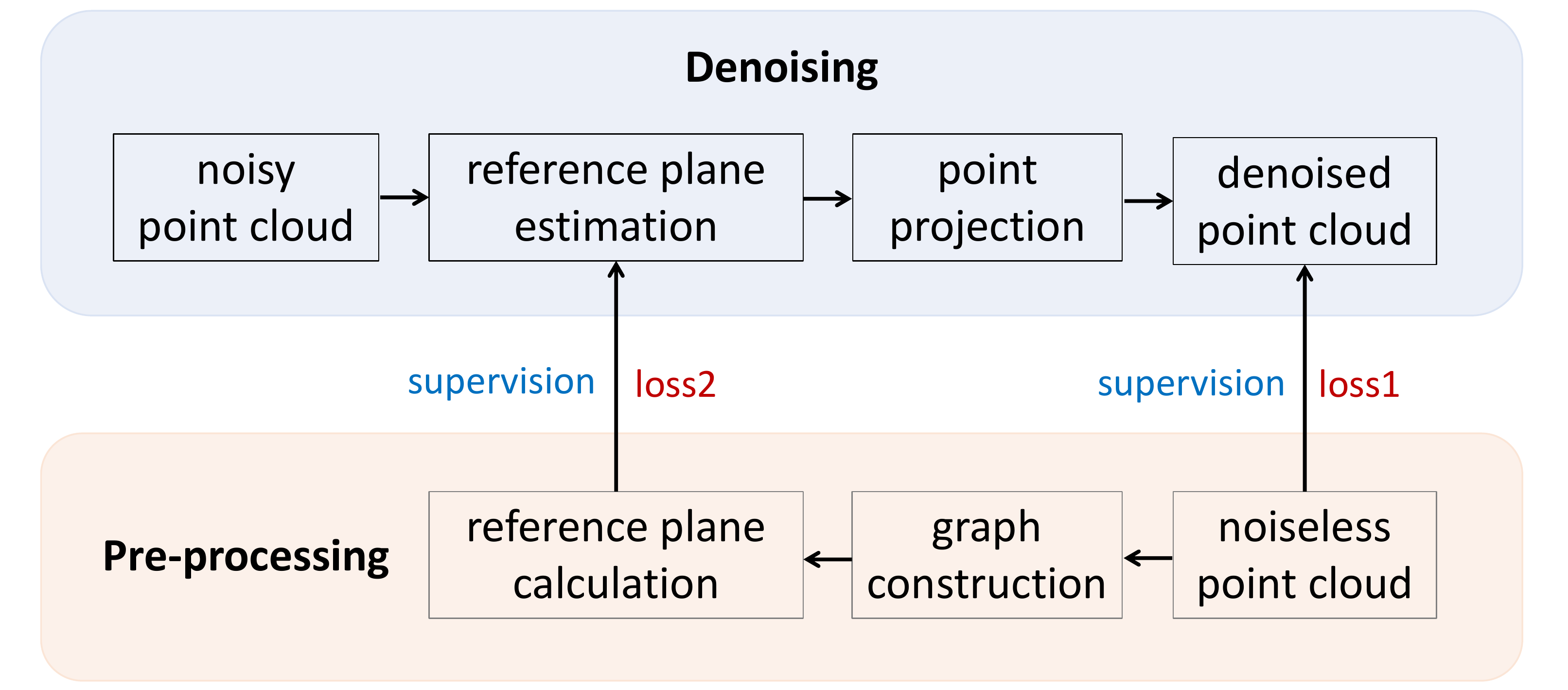}}
\caption{\textbf{System pipeline:} During pre-processing, we calculate reference planes for points in noiseless point clouds as in~\cite{WMP}. During denoising, we process noisy point clouds to estimate reference planes for points. We then project noisy points to their corresponding estimated reference planes to denoise the point clouds. The noiseless point clouds and the calculated reference planes are used to supervise the denoised point clouds and the estimated reference planes, respectively.}\label{pipeline}
\end{figure}
\vspace{-3mm}

\textbf{Algorithm.}
The system pipeline is shown in Fig. \ref{pipeline}. During pre-processing, we use graph-based techniques to calculate reference planes for points as in~\cite{WMP}. Let $T_i$ be the reference plane with normal vector $\mathbf{a}_i$ and interception $c_i$ calculated from noiseless point cloud for point $\mathbf{\mathbf{p}_i}$. Let $\widehat{T}_i$ be the reference plane with normal vector $\widehat{\mathbf{a}}_i $ and interception $\widehat{c}_i$ estimated from our neural network. 

During denoising procedure, we project noisy point $\widetilde{\mathbf{p}}_i$ to the estimated plane $\widehat{T}_i$ to obtain the denoised point $\widehat{\mathbf{p}}_i$ as
$$
\vspace{-1mm}
\widehat{\mathbf{p}}_i = \widetilde{\mathbf{p}}_i - \widehat{\mathbf{a}}_i^T\widetilde{\mathbf{p}}_i\widehat{\mathbf{a}}_i + \widehat{c}_i\widehat{\mathbf{a}}_i.
\vspace{-1mm}
$$
The reference plane and the noiseless point cloud are used to constraint or supervise the reference plane estimation and the denoised point cloud, respectively.

\begin{figure*}[htb!]
\centerline{\includegraphics[width=6.2in , height = 2.4in, keepaspectratio]{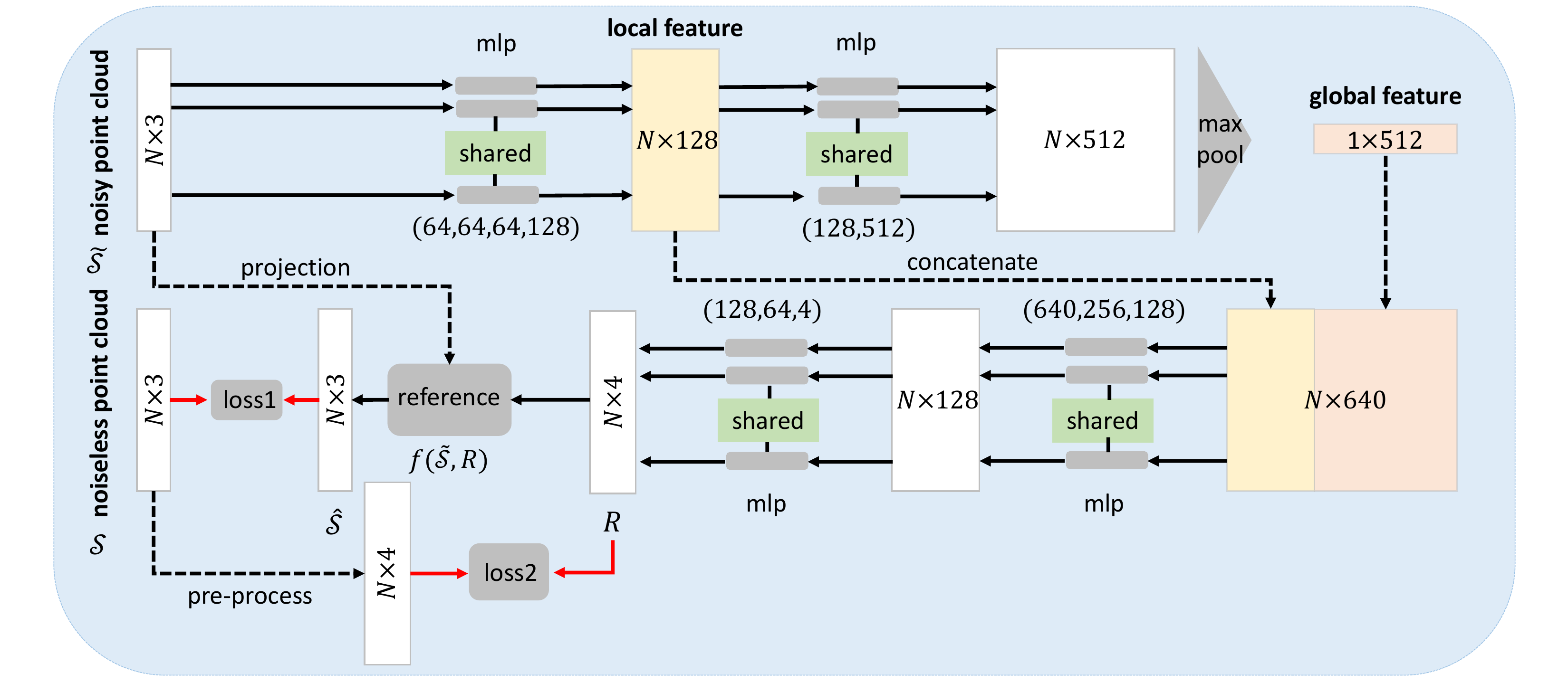}}
\caption{\textbf{NPD architecture}: The proposed network takes noisy point clouds $\widetilde{\mathcal{S}}$ with $N$ points as input, then passes the inputs through the shared multi-layer perceptron (MLP) to obtain local features. We mark MLP as shared because each point goes through the same MLP and the feature vector is obtained independently and identically as in~\cite{2016pointnet}. The global features are captured by max-pooling, which selects the max value of each of 512 columns to produce a $1 \times 512$ global feature vector. We concatenate global and local features, then pass them through MLP to obtain reference planes represented as an $N \times 4$ matrix. A function $f(\widetilde{\mathcal{S}} , R)$ is applied to project the noisy points to the estimated reference planes to denoise point clouds. The denoised point cloud $\widehat{\mathcal{S}}$ and estimated reference planes $\widehat{T}_i$ are supervised by the noiseless point cloud $\mathcal{S}$ and the calculated reference plane $T_i$, respectively. We use MSE loss to calculate \textit{loss1} (Eq. \ref{mse}), which constraints the denoised point clouds to stay close to the noiseless point clouds. We use the cosine similarity loss to calculate \textit{loss2} (Eq. \ref{loss2}), which constrains the estimated reference planes $\widehat{T}_i$ to be similar to the calculated reference planes $T_i$ \cite{WMP}. The number of MLP layers are shown in brackets. }\label{archt}
\vspace{-2mm}
\end{figure*}

NPD architecture is shown in Fig. \ref{archt}. The architecture directly takes noisy point clouds as inputs and passes through shared multi-layer perceptron (MLP), which means each perceptron is applied to the feature vectors of each point independently and identically as in~\cite{2016pointnet}. The max-pooling is selecting the max value of each column of the matrix; the global information is contained in a 1D vector for each point cloud and represents a whole point cloud~\cite{2016pointnet}. 

The local feature vector contains local information for each point. The extracted global information is concatenated to make our architecture less sensitive to sampling densities and able to learn the number of neighboring points adaptively. 3D point cloud denoising requires local information of each point since reference planes are different for different points due to their local geometries. For example, a point on a flat surface and a point on a sharp edge contain different local information and they should be treated differently. 3D point cloud denoising also requires global information for each point in a point cloud. For example, the edge distribution and the curvature variation of an airplane and a table are distinct. It is essential that each point is aware of the skeleton and the global view of the surface from which it is sampled.

We pass the combined information through shared MLPs to learn reference planes. We then project the points in noisy point clouds to their corresponding reference planes. The projected points are the denoised points sampled from the reference planes. \textit{loss1} is treated as a point-cloud loss and we calculate mean-squared-error (MSE) between the noiseless point cloud $\mathcal{S}$ and the denoised point cloud $\widehat{\mathcal{S}}$, 
\vspace{-3mm}
\begin{equation}\label{mse}
    \rm{loss1} (\widehat{\mathbf{p}}_i) = \text{MSE}(\widehat{\mathcal{S}} , \mathcal{S}) = \frac{1}{N}\sum_i \lVert \widehat{\mathbf{p}}_i - \mathbf{p}_i \rVert_2^2.
    \vspace{-1mm}
\end{equation}
\textit{loss2} is treated as a reference-plane loss and we calculate the absolute value of the cosine similarity. 
\begin{equation}\label{loss2}
    \rm{loss2} (\widehat{\mathbf{a}}_i , \widehat{c}_i) = \frac{1}{N}\sum_i|1 - \frac{[\mathbf{a}_i^T , c_i]^T [\widehat{\mathbf{a}}_i^T , \widehat{c}_i] }{\lVert [\mathbf{a}_i^T , c_i] \rVert_2  \lVert [\widehat{\mathbf{a}}_i^T , \widehat{c}_i] \rVert_2}|.
\end{equation}

The training loss is calculated as
$$\rm{loss} = (\alpha*\rm{loss1} + (1 - \alpha) * \rm{loss2}),$$
where $\alpha \in [0 , 1]$ is a parameter that we can tune during the training for the best performance. We also show the effect of $\alpha$ with three small datasets in Fig. \ref{alpha}. 


\section{Experimental results}
\textbf{Dataset.}
We train our network with ShapeNet data, which contains 55 object categories with more than 50,000 3D mesh models \cite{shapenet}. We test our net with the dataset in ModelNet10, which contains 10 object categories with more than 3000 3D mesh models~\cite{modelnettest}. We select 40000 3D meshes in ShapeNet and sampled point clouds for the training procedure and each point cloud contains 2048 points; we randomly select 50 3D meshes in seven categories in ModelNet10 and sample point clouds for the testing procedure. The point clouds are rescaled into a unit cube and centered at the origin. We pre-process the training data to obtain reference planes of points and add i.i.d. Gaussian noise to construct noisy point clouds. We compare the denoising results from NPD with the state-of-the-art denoising algorithms, including BF algorithm~\cite{denoise_bf}, GDB algorithm~\cite{denoise_graph}, PDE algorithm~\cite{denoise_pde1}, Non-local Denoising (NLD) algorithm~\cite{denoise_nld} and MWP~\cite{WMP}. For all the algorithms, we tune parameters to produce their best performances. 

\textbf{Results and analysis.}
To quantify the performances of different algorithms, we use the metrics MSE (Eq.(\ref{mse})) and Chamfer distance (CD) defined as, 
\small
\begin{equation*}
\begin{aligned}
\text{CD}(\widehat{\mathcal{S}} , \mathcal{S}) = \frac{1}{N} \Big(\sum_{\widehat{\mathbf{p}}_i\in {\widehat{\mathcal{S}}}}\min_{\mathbf{p}_j\in {\mathcal{S}}}||\mathbf{p}_i - \widehat{\mathbf{p}}_j||_2^2 
 + \sum_{\mathbf{p}_i\in {\mathcal{S}}}\min_{\widehat{\mathbf{p}}_j\in {\widehat{\mathcal{S}}}}||\widehat{\mathbf{p}}_j - \mathbf{p}_i||_2^2 \Big),
\end{aligned}
\end{equation*}
\normalsize
where $\widehat{\mathcal{S}}$ and $\mathcal{S}$ denote denoised point clouds and noiseless point clouds, respectively. We have $\widehat{\mathbf{p}}_i \in \widehat{\mathcal{S}}$, $\mathbf{p}_i \in \mathcal{S}$ and $N=|\widetilde{\mathcal{S}}|=|\mathcal{S}|$ is the total number of points.

\begin{figure}[htbp]
	\centerline{\includegraphics[width=3.3in]{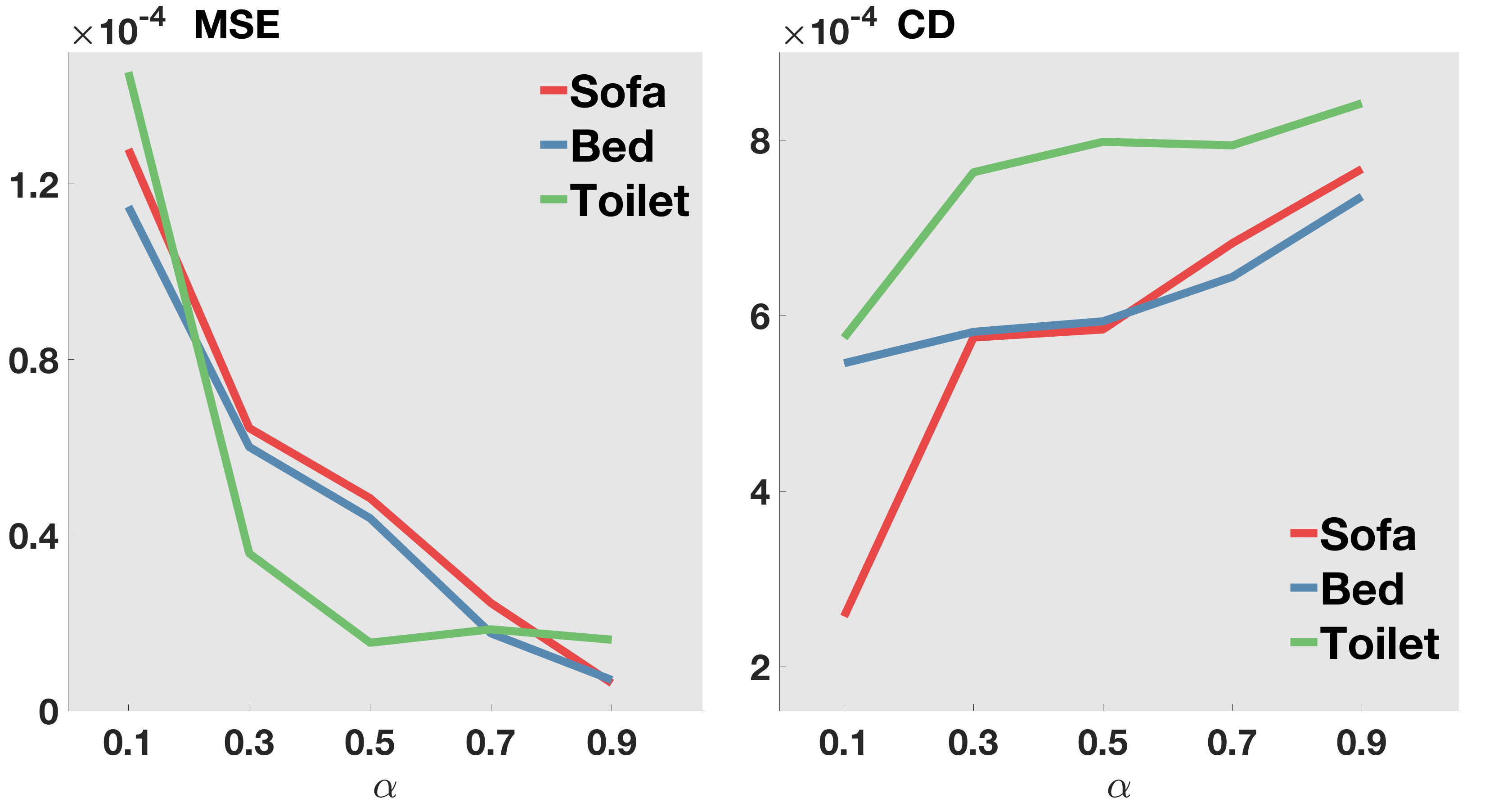}}
	\caption{For all three categories (sofa, bed, and toilet), MSE (left) decreases and CD (right) increases as $\alpha$ increases.}
	\label{alpha}
\end{figure}

To show the effects of $\alpha$ in our loss function, we vary $\alpha = 0.1, 0.3, 0.5, 0.7, 0.9$ to train the neural network with smaller datasets. MSE and CD errors are shown in Fig. \ref{alpha}.

The tendencies of MSE error and CD error are different as the value of $\alpha$ increases. CD promotes the plane-to-plane matching; MSE promotes the point-to-point matching. 1) When $\alpha \rightarrow 0$, we restrict the predicted reference planes stay closer to the calculated reference planes. It makes the manifolds that the denoised point clouds are lying on close to the manifolds that the noiseless point clouds are lying on, which produces the smaller CD values; 2) When $\alpha \rightarrow 1$, we restrict the  point clouds $\widehat{\mathcal{S}}$ calculated from the predicted reference planes stay close to the noiseless point clouds $\mathcal{S}$. It makes the denoised point clouds stay close to the noiseless point clouds, which produces the smaller MSE values.

We also train a network without providing the global feature in Fig. \ref{archt} and the errors are shown in Table \ref{comp}. 
\begin{table}[htb!]
	\begin{center}
		\begin{tabular}{ c|c|c|c|c } 
			\hline
			Loss ($10^{-4}$) & \multicolumn{2}{c}{Local+global features} & \multicolumn{2}{c}{local feature only} \\
			\hline
			Category & MSE & CD & MSE & CD\\
			\hline
			Sofa  & \bf{0.627} & \bf{0.7670} & 10.845 & 10.4758\\
			Bed  & \bf{0.700} & \bf{0.7356} & 13.5513 & 9.9097\\
			Toilet & \bf{1.617} & \bf{0.8418} & 11.307 & 9.8542  \\
			\hline
		\end{tabular}
	\end{center}
	\caption{MSE error and CD error for three categories (sofa, bed, and toilet) with and without global features. It shows that the global features significantly improve the denoising performance by providing global contextual information, which is rarely considered in most previous works on denoising.}\label{comp}
\end{table}

\begin{figure}[htbp]
\centerline{\includegraphics[width=3.6in]{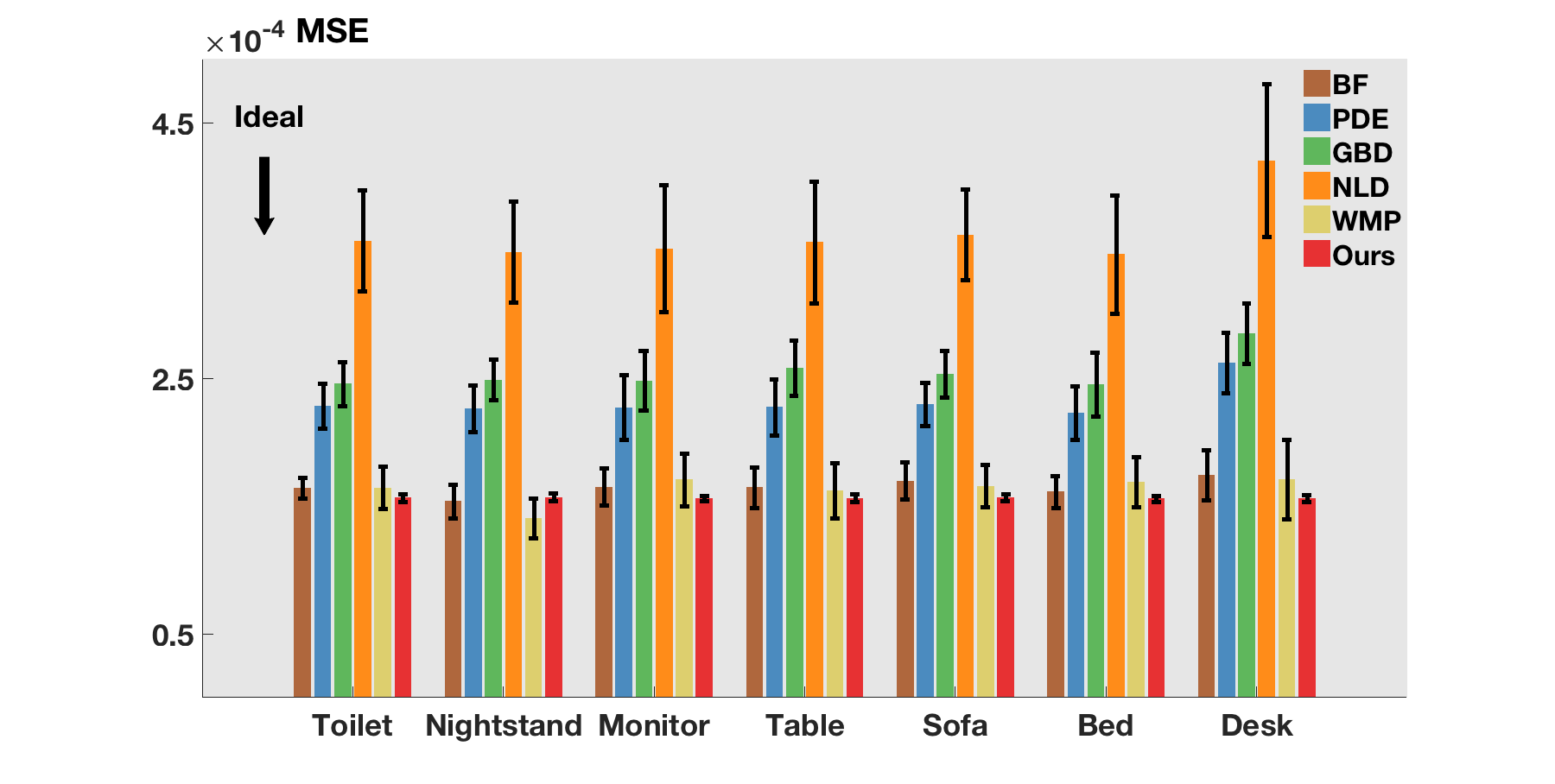}}
\caption{The proposed algorithm (in red) outperforms all of its competitors in terms of MSE. The denoised point clouds produced by our architecture stay point-wise closer to the noiseless point clouds compared to the denoised point clouds produced by other algorithms.}
\label{mse_mean}
\end{figure}

\begin{figure}[htbp]
\centerline{\includegraphics[width=3.6in]{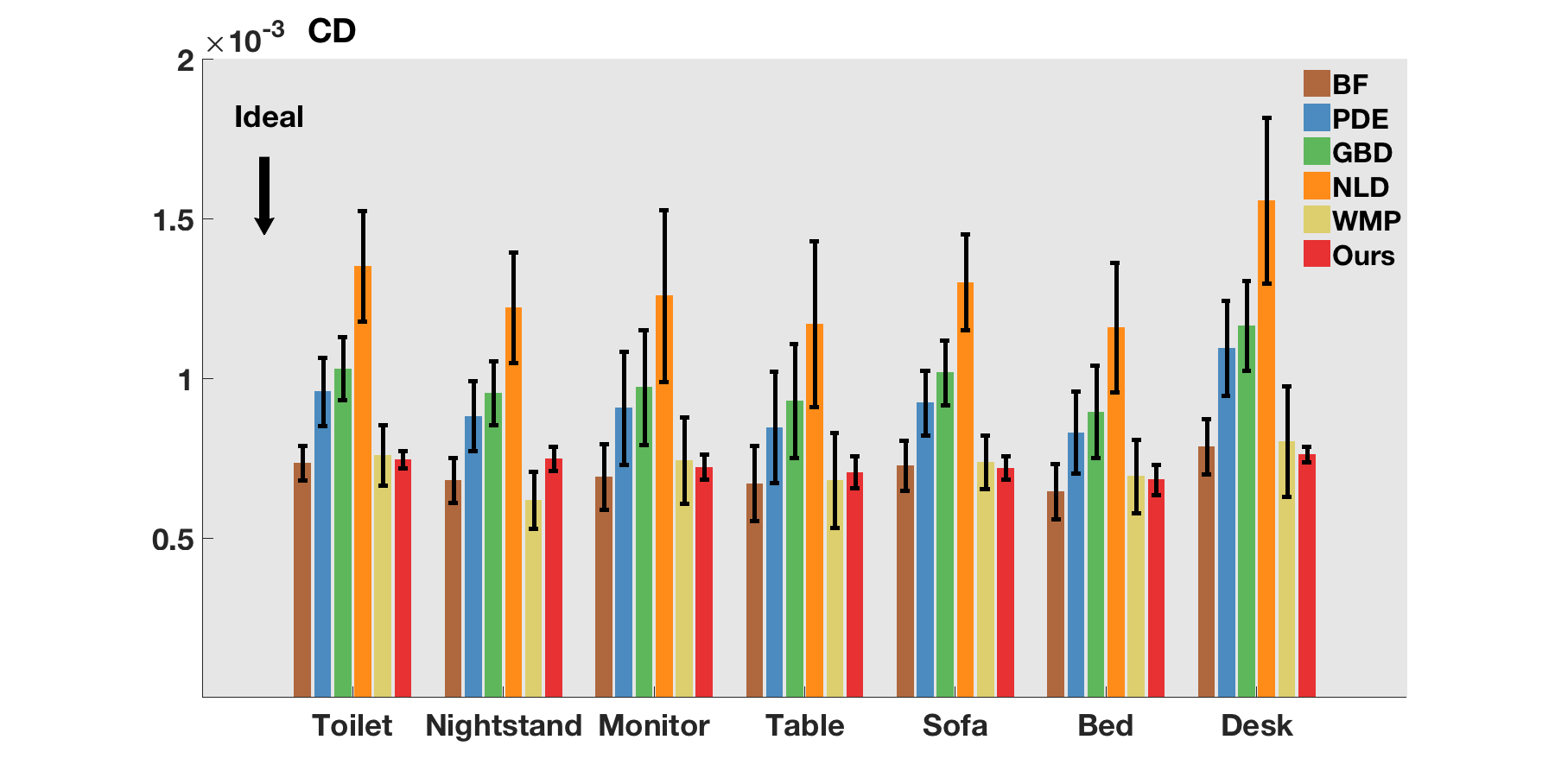}}
\caption{The proposed algorithm (in red) outperforms its competitors in terms of CD. The denoised point clouds produced by our architecture are closer to the original point clouds or the manifolds point clouds are lying on compared to the denoised point clouds produced by other algorithms.}
\label{cd_mean}
\end{figure}

In our experiments, we start the training with small $\alpha$ and increase the value of $\alpha$ gradually. Figs. \ref{mse_mean} and Fig. \ref{cd_mean} show the denoising performances in seven categories evaluated by MSE and CD (mean and standard deviation), respectively. We see that NPD algorithm outperforms four of its competitors in all seven categories by the defined metrics. NPD algorithm also produces the smallest variance for all the categories and both evaluation metrics. We analyze the reason to be that the number of point clouds and the number of object categories in the training dataset are large enough and the trained network is powerful enough to denoise all the point clouds in our testing dataset. 

\vspace{-4mm}
\section{Conclusions}\label{conclusion}
We propose a novel algorithm for 3D point cloud denoising by estimating reference planes for points with deep learning techniques called neural projection algorithm (NPD). Without searching for neighboring points for each point in noisy point clouds as previous algorithms, we directly estimate reference planes and project noisy points to their corresponding reference planes. We validate the proposed architecture on real datasets and the proposed architecture beats BF, PDE, GBD, and NLD algorithms. The proposed algorithm produces a much smaller variance in all seven categories for both two evaluation metrics.

\newpage 
\bibliographystyle{IEEEtran}
\bibliography{0my_ref}

\begin{thebibliography}{10}
\providecommand{\url}[1]{#1}
\csname url@samestyle\endcsname
\providecommand{\newblock}{\relax}
\providecommand{\bibinfo}[2]{#2}
\providecommand{\BIBentrySTDinterwordspacing}{\spaceskip=0pt\relax}
\providecommand{\BIBentryALTinterwordstretchfactor}{4}
\providecommand{\BIBentryALTinterwordspacing}{\spaceskip=\fontdimen2\font plus
\BIBentryALTinterwordstretchfactor\fontdimen3\font minus
  \fontdimen4\font\relax}
\providecommand{\BIBforeignlanguage}[2]{{%
\expandafter\ifx\csname l@#1\endcsname\relax
\typeout{** WARNING: IEEEtran.bst: No hyphenation pattern has been}%
\typeout{** loaded for the language `#1'. Using the pattern for}%
\typeout{** the default language instead.}%
\else
\language=\csname l@#1\endcsname
\fi
#2}}
\providecommand{\BIBdecl}{\relax}
\BIBdecl

\bibitem{pcl}
R.~B. Rusu and S.~Cousins, ``3d is here: Point cloud library (pcl),''
  \emph{2011 IEEE International Conference on Robotics and Automation}, pp.
  1--4, 2011.

\bibitem{pc_tutorial}
A.~Aldoma, Z.~C. Marton, F.~Tombari, W.~Wohlkinger, C.~Potthast, B.~Zeisl,
  R.~B. Rusu, S.~Gedikli, and M.~Vincze, ``Tutorial: Point cloud library:
  Three-dimensional object recognition and 6 dof pose estimation,'' \emph{IEEE
  Robotics Automation Magazine}, vol.~19, no.~3, pp. 80--91, Sept 2012.

\bibitem{sample_surface}
J.~Park, H.~Kim, Y.-W. Tai, M.~S. Brown, and I.~Kweon, ``High quality depth map
  upsampling for 3d-tof cameras,'' in \emph{2011 International Conference on
  Computer Vision}, Nov 2011, pp. 1623--1630.

\bibitem{sensor}
J.~Han, L.~Shao, D.~Xu, and J.~Shotton, ``Enhanced computer vision with
  microsoft kinect sensor: A review,'' \emph{IEEE Transactions on Cybernetics},
  vol.~43, no.~5, pp. 1318--1334, Oct 2013.

\bibitem{image_pc}
Y.~Furukawa and J.~Ponce, ``Accurate, dense, and robust multiview stereopsis,''
  \emph{IEEE Transactions on Pattern Analysis and Machine Intelligence},
  vol.~32, no.~8, pp. 1362--1376, Aug 2010.

\bibitem{machine_noise}
H.~Chen and J.~Shen, ``Denoising of point cloud data for computer-aided design,
  engineering, and manufacturing,'' \emph{Engineering with Computers}, vol.~34,
  no.~3, pp. 523--541, 2018.

\bibitem{siheng}
S.~Chen, D.~Tian, C.~Feng, A.~Vetro, and J.~Kovačević, ``Fast resampling of
  three-dimensional point clouds via graphs,'' \emph{IEEE Transactions on
  Signal Processing}, vol.~66, no.~3, pp. 666--681, Feb 2018.

\bibitem{dl_language}
R.~Collobert and J.~Weston, ``A unified architecture for natural language
  processing: Deep neural networks with multitask learning,'' in
  \emph{Proceedings of the 25th International Conference on Machine Learning},
  ser. ICML '08.\hskip 1em plus 0.5em minus 0.4em\relax New York, NY, USA: ACM,
  2008, pp. 160--167.

\bibitem{2016pointnet}
C.~R. Qi, H.~Su, K.~Mo, and L.~J. Guibas, ``Pointnet: Deep learning on point
  sets for 3d classification and segmentation,'' in \emph{Proceedings of the
  IEEE Conference on Computer Vision and Pattern Recognition}, 2017, pp.
  652--660.

\bibitem{folding}
Y.~Yang, C.~Feng, Y.~Shen, and D.~Tian, ``Foldingnet: Interpretable
  unsupervised learning on 3d point clouds,'' \emph{arXiv preprint
  arXiv:1712.07262}, 2018.

\bibitem{upsampling}
L.~Yu, X.~Li, C.-W. Fu, D.~Cohen-Or, and P.-A. Heng, ``Pu-net: Point cloud
  upsampling network,'' in \emph{Proceedings of the IEEE Conference on Computer
  Vision and Pattern Recognition}, 2018, pp. 2790--2799.

\bibitem{shapenet}
A.~X. Chang, T.~Funkhouser, L.~Guibas, P.~Hanrahan, Q.~Huang, Z.~Li,
  S.~Savarese, M.~Savva, S.~Song, H.~Su \emph{et~al.}, ``Shapenet: An
  information-rich 3d model repository,'' \emph{arXiv preprint
  arXiv:1512.03012}, 2015.

\bibitem{modelnettest}
Z.~Wu, S.~Song, A.~Khosla, F.~Yu, L.~Zhang, X.~Tang, and J.~Xiao, ``3d
  shapenets: A deep representation for volumetric shapes,'' in \emph{2015 IEEE
  Conference on Computer Vision and Pattern Recognition (CVPR)}, June 2015, pp.
  1912--1920.

\bibitem{denoise_bf}
J.~Digne and C.~de~Franchis, ``{The Bilateral Filter for Point Clouds},''
  \emph{{Image Processing On Line}}, vol.~7, pp. 278--287, 2017.

\bibitem{denoise_pde1}
A.~Elmoataz, O.~Lezoray, and S.~Bougleux, ``Nonlocal discrete regularization on
  weighted graphs: A framework for image and manifold processing,''
  \emph{Trans. Img. Proc.}, vol.~17, no.~7, pp. 1047--1060, Jul. 2008.

\bibitem{denoise_nld}
J.-E. Deschaud and F.~Goulette, ``Point cloud non local denoising using local
  surface descriptor similarity,'' \emph{IAPRS}, vol.~38, no.~3A, pp. 109--114,
  2010.

\bibitem{graph_manifold}
M.~Belkin and P.~Niyogi, ``Towards a theoretical foundation for laplacian-based
  manifold methods,'' in \emph{Learning Theory}.\hskip 1em plus 0.5em minus
  0.4em\relax Berlin, Heidelberg: Springer Berlin Heidelberg, 2005, pp.
  486--500.

\bibitem{newgraph}
J.~Zeng, G.~Cheung, M.~Ng, J.~Pang, and C.~Yang, ``3d point cloud denoising
  using graph laplacian regularization of a low dimensional manifold model,''
  \emph{arXiv preprint arXiv:1803.07252}, 2018.

\bibitem{graph_analysis}
C.~Dinesh, G.~Cheung, I.~V. Bajic, and C.~Yang, ``Fast 3d point cloud denoising
  via bipartite graph approximation \& total variation.''\hskip 1em plus 0.5em
  minus 0.4em\relax arXiv:1804.10831, 2018.

\bibitem{WMP}
C.~Duan, S.~Chen, and J.~Kovačević, ``Weighted multi-projection: 3d point
  cloud denoising with tangent planes,'' \emph{IEEE Global Conference on Signal
  and Informationn Processing}, 2018.

\bibitem{denoise_graph}
Y.~Schoenenberger, J.~Paratte, and P.~Vandergheynst, ``Graph-based denoising
  for time-varying point clouds,'' in \emph{2015 3DTV-Conference: The True
  Vision-Capture, Transmission and Display of 3D Video (3DTV-CON)}.\hskip 1em
  plus 0.5em minus 0.4em\relax IEEE, 2015, pp. 1--4.

\end{thebibliography}

\end{document}